\documentclass[10pt, a4paper]{article}

\usepackage[table,xcdraw]{xcolor}
\usepackage{lrec-coling2024} 
\usepackage{comment}
\usepackage{latexsym}
\usepackage{graphicx}
\usepackage{makecell} 
\usepackage{booktabs}
\usepackage{fancyvrb}

\title{DANSK and DaCy 2.6.0: Domain Generalization of Danish Named Entity Recognition}

\name{Kenneth Enevoldsen, Emil Trenckner Jessen, Rebekah Baglini} 

\address{Aarhus University}

\abstract{
Named entity recognition is one of the cornerstones of Danish NLP, essential for language technology applications within both industry and research. However, Danish NER is inhibited by a lack of available datasets. As a consequence, no current models are capable of fine-grained named entity recognition, nor have they been evaluated for potential generalizability issues across datasets and domains. To alleviate these limitations, this paper introduces: 1) DANSK: a named entity dataset providing for high-granularity tagging as well as within-domain evaluation of models across a diverse set of domains; 
2) DaCy 2.6.0 that includes three generalizable models with fine-grained annotation; and 3) an evaluation of 
current state-of-the-art models' ability to generalize across domains. The evaluation of existing and new models revealed notable performance discrepancies across domains, which should be addressed within the field. Shortcomings of the annotation quality of the dataset and its impact on model training and evaluation are also discussed. Despite these limitation, we advocate for the use of the new dataset DANSK alongside further work on the  generalizability within Danish NER.
 \\ \newline  }

\begin{document}

\maketitleabstract

\section{Introduction}

\textbf{D}anish \textbf{A}nnotations for \textbf{N}LP \textbf{S}pecific Tas\textbf{K}s (\textbf{DANSK}) is a new gold-standard dataset for Danish with named entity annotations for 18 distinct classes. The annotated texts are from 25 text sources that span 7 different domains and have been derived from the Danish Gigaword Corpus \cite{stromberg-derczynski_danish_2021}. The dataset is publicly accessible\footnote{\href{https://anonymous.4open.science/r/dansk-3A03}{https://anonymous.4open.science/r/dansk-3A03}} and pre-partitioned into a training, validation, and testing set in order to standardize future model evaluations. 

The release of DANSK is motivated by present limitations facing Danish NER. The first limitation concerns a lack of generalizability measures of current SOTA models: all have been either fully or partially fine-tuned for the NER task on a single dataset, Danish Named Entities (DaNE)  \citep{hvingelby_dane_2020}. Although DaNE features high-quality NER annotations 
and features texts from a wide array of domains and sources, it has several shortcomings. 
First, domains such as social media and legal texts are lacking from DaNE entirely and spoken language is severely underrepresented. Moreover, since the texts are from 1883-1992, no contemporary linguistic trends are included. While current Danish models perform quite well on DaNE \cite{nielsen_scandeval_2023}, their performances is naturally an expression of performance on the texts that are included. 

Second, individual domain evaluation is not possible even for domains included in the dataset, as DaNE lacks metadata on the origin of the texts.  Information on domain biases is therefore occluded in any evaluations. This is especially problematic because many models' current use cases are on texts that are not represented in DaNE; e.g. on social media data.

Third, DaNE uses the CoNLL-2003 annotation standard consisting of four types, as opposed to more fine-grained NER datasets like OntoNotes 5.0 with 18 entity types. 

Thus, Danish NLP is in need of more open and free datasets, in part to navigate impediments to generalizability \citep{kirkedal_lacunae_2019}. Domain shifts in data cause drops in performance, as models are optimized for the training and validation data, making cross-domain evaluation, particularly for tasks like NER, crucial \citep{plank_dan_2021}. A study by \citet{enevoldsen_dacy_2021}, furthermore found generalizability issues for NER in Danish, not across domains, but across different types of data augmentations --- further indicating generalizability issues for Danish models.

The DANSK dataset was designed to address these limitations currently facing Danish NER. Based on DANSK, we also introduce three new models of varying sizes incorporated into DaCy \cite{enevoldsen_dacy_2021} that are specifically developed for fine-grained NER on the comprehensive array of domains included in DANSK to ensure generalizability. 

Finally, we evaluate the three newly released DaCY models against some of the currently best-performing and most widely-used NLP models within Danish NER using the DANSK dataset, in order to attain estimates of generalizability across domains.

\section{Dataset}

\subsection{The Danish Gigaword Corpus} 
The texts in the DANSK dataset were sampled from the Danish Gigaword Corpus (DAGW) \cite{stromberg-derczynski_danish_2021}, a new Danish corpus of over 1 billion words, consisting of 25 different media sources across 7 domains\footnote{"Domains" within DANSK are inherited directly from the Danish Gigaword Corpus(DAGW) \citep{stromberg-derczynski_danish_2021}. Naturally, some domains constitute more coherent genres of text than others (e.g. "Legal" versus "Web" or "Social Media" but we have retained these labels to maintain consistency with DAGW. We take domain to refer to a distinct area or field of knowledge or activity characterized by its specific terminology, linguistic patterns, and/or unique challenges in language processing.} (see Appendix A.3.2 for discussion of domains).

\subsection{Initial named entity annotation}
 For annotation of DANSK, DAGW was filtered to exclude texts from prior to 2000 and segmented into sentences using spaCy's rule-based "sentencizer" \citep{honnibal_spacy_2020}. DANSK uses the annotation standard of OntoNotes 5.0. For NER annotation using Prodigy \citep{montani2018prodigy}, texts were first divided up equally for the 10 annotators, with a 10\% overlap between the assigned texts (i.e. 10\% of texts were annotated by all annotators). The annotators were 10 native speakers of Danish (nine female, one male) between the ages of 22-30 years old, studying in the Masters degree program in English Linguistics at Aarhus University. For fine-grained NER annotation, instructions followed the 18 shorthand descriptions of the OntoNotes 5.0 named entity types \citep{weischedel_ontonotes_2012}. For more information on the recruitment and compensation of annotators and the annotation instruction process, see Section 8 and Appendix A.4.2 upon publication. 
Initial annotations suffered from poor intercoder reliability, as measured by Cohen's kappa ($\kappa$) scores over tokens, calculated by matching each rater pairwise to every other (Table \ref{tab:initial_kappa}). In order to assess the annotation consensus between annotators on a entity type level, additional F1-mean scores were calculated for all annotators (Table \ref{tab:initial_tag_f1}). For the calculations of the Span-F1 score, we used the spacy implementation (v. 3.5.4). 

\begin{table}
    \centering
        \begin{tabular}{l|cc}
\hline
\rowcolor[HTML]{EFEFEF}
& \multicolumn{2}{c}{\textbf{Cohen's $\kappa$}} \\
\rowcolor[HTML]{EFEFEF}
 & Initial & Reviewed \\
\hline
Annotator 1 & 0.6 & 0.92 \\
Annotator 2 & 0.52 & - \\
Annotator 3 & 0.51 & 0.93 \\
Annotator 4 & 0.58 & 0.93 \\
 Annotator 5 & 0.54 & 0.91 \\
Annotator 6 & 0.56 & 0.93 \\
Annotator 7 & 0.47 & 0.93 \\
Annotator 8 & 0.51 & 0.89 \\
Annotator 9 & 0.52 & 0.92 \\
Annotator 10 & 0.56 & - \\
\hline
Average & & 0.92 \\
\hline
\end{tabular}
        \caption{Table showing the average Cohen's \(\kappa\) scores for each rater for the overlapping data after the initial annotation and after the annotations were reviewed and improved (see section \ref{sec:anno_improvement}).}
        \label{tab:initial_kappa}
\end{table}

\begin{table}[h]
    \centering

\begin{tabular}{lll}
\toprule
    
\textbf{Named-entity type} & \textbf{Mean F1} & \textbf{F1 SD} \\ 
\midrule
\midrule

CARDINAL                   & 0.47                   & 0.23                           \\        
\rowcolor[HTML]{EFEFEF} 
DATE                       & 0.55                   & 0.21                           \\        
EVENT                      & 0.5                    & 0.34                           \\        
\rowcolor[HTML]{EFEFEF} 
FACILITY                   & 0.22                   & 0.38                           \\        
GPE                        & 0.91                   & 0.05                           \\        
\rowcolor[HTML]{EFEFEF} 
LANGUAGE                   & 0.0                    & 0.0                            \\        
LAW                        & 0.23                   & 0.32                           \\        
\rowcolor[HTML]{EFEFEF} 
LOCATION                   & 0.22                   & 0.24                           \\        
MONEY                      & 0.62                   & 0.49                           \\        
\rowcolor[HTML]{EFEFEF} 
NORP                       & 0.5                    & 0.39                           \\        
ORDINAL                    & 0.5                    & 0.27                           \\        
\rowcolor[HTML]{EFEFEF} 
ORGANIZATION               & 0.72                   & 0.14                           \\        
PERCENT                    & 0.0                    & 0.0                            \\        
\rowcolor[HTML]{EFEFEF} 
PERSON                     & 0.59                   & 0.32                           \\        
PRODUCT                    & 0.12                   & 0.23                           \\        
\rowcolor[HTML]{EFEFEF} 
QUANTITY                   & 0.18                   & 0.26                           \\        
TIME                       & 0.33                   & 0.36                           \\        
\rowcolor[HTML]{EFEFEF} 
WORK OF ART                & 0.4                    & 0.29                           \\        
\end{tabular}

        \caption{The mean and standard deviation of the F1-scores across the raters for each of the named entity types.}
        \label{tab:initial_tag_f1}
\end{table}

\subsection{Annotation improvement}
\label{sec:anno_improvement}

Due to the low consensus between annotators, it was deemed necessary for the annotated texts to undergo additional processing before they could be unified into a coherent, high-quality dataset. 

\paragraph{Texts with multiple annotators}

Some curated datasets utilize a single annotator for manual resolvement of conflicts between raters \citep{weischedel_ontonotes_2012}. While this is sometimes necessary, it skews annotations towards the opinion of a single annotator rather than the general consensus across raters. In order to resolve conflicts while diminishing this skew, we took a two-step approach: first, an automated procedure was employed to resolve the majority of annotation disagreements systematically; second, a small number of texts with remaining annotation conflicts were resolved manually. 

\begin{figure}[h!]
    \centering
    \includegraphics[scale=.150]{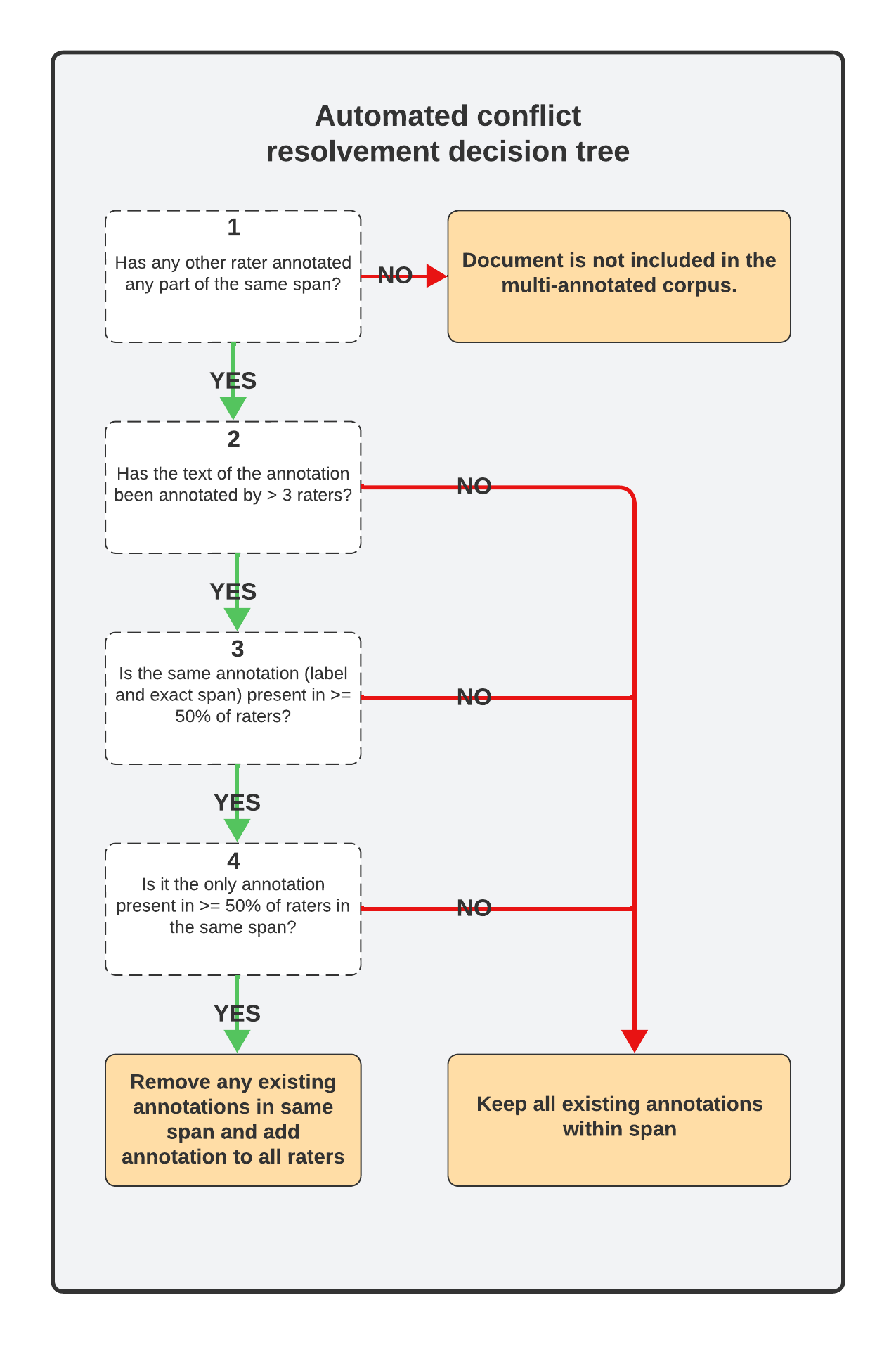}
    \caption{The decision tree for automated conflict resolvement of multi-annotated texts. Each annotation span in a text followed the steps from 1 to 4 on the diagram. The decision tree was only followed for annotation spans found in texts that had been annotated by at least four raters.}
    \label{fig:automated_conflict_resolvement_rules}
\end{figure}

\begin{figure}[h!]
    \includegraphics[scale=.255]{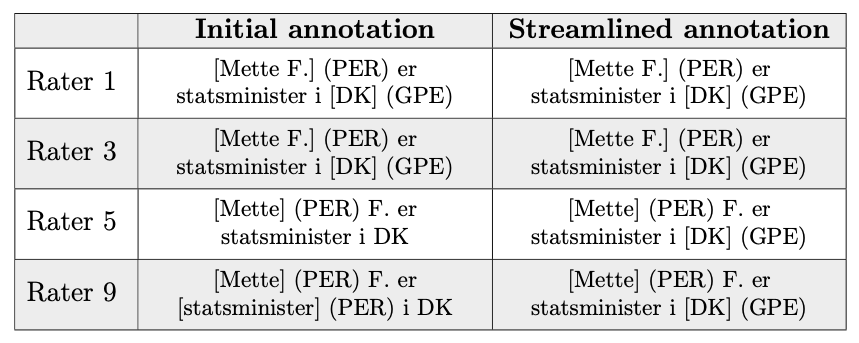}
        \caption{An example of a text along with its four annotations being processed on the basis of the decision-tree in Figure \ref{fig:automated_conflict_resolvement_rules}.
        \label{fig:automated_conflict_resolvement_example}}
\end{figure}

The automated procedure for resolving annotation disagreements was rule-based and followed a decision tree-like structure (Figure \ref{fig:automated_conflict_resolvement_rules}). It was only applied to texts that had been annotated by a minimum of four raters, ensuring that that an annotation with no consensus was accepted in a text annotated by two annotators. To exemplify the streamlining of the multi-annotated texts, Figure \ref{fig:automated_conflict_resolvement_example} is included.

After employing the automated procedure, the 886 multi-annotated texts went from having 513 (58\%) texts with complete rater agreement to 789 (89\%). The texts with complete agreement were added to the DANSK dataset, while the remaining 97 (21\%) of the multi-annotated texts had remaining annotation conflicts. The remaining texts with conflicting annotations were resolved manually by the first author, by changing any annotations that did not comply with the extended OntoNotes annotation guidelines. However, three texts were of such bad quality that they were rejected and excluded. The remaining resolved 94 texts were then added to DANSK.

Finally, to ensure that any named entities of the type LANGUAGE, PERCENT, and PRODUCT had not been missed by the annotators, an extra measure was taken. The model {\tt TNER/Roberta-Large-OntoNotes5}\footnote{https://huggingface.co/tner/roberta-large-ontonotes5} was used to add these types of annotations to the accepted multi-annotated texts \citep{ushio_t-ner_2021}. Each text with any predictions by the models was then manually assessed by the first author, to inspect whether the additional model annotations should be included. None of the predictions matched the annotation guidelines and were thus not added to the texts.
This step concluded the processing of the multi-annotated texts, which resulted in a total of 883 texts added to the DANSK dataset.

\paragraph{Texts with a single annotator}

Based on the low consensus between the multiple raters, it was assumed that documents annotated by a single annotator might not meet a sufficient quality standard.  To refine these annotations, we utilize the reviewed annotations from multiple annotators to train a model. This model is then applied to the data such that detected discrepancies between model and human annotations are reviewed and manually resolved by the authors. The rationale for this process is that it propagates the aggregated annotations across the dataset and can thus be seen as a supervised approach to anomaly detection

As the preliminary DANSK dataset included relatively few annotations, we explored the effect of enriching our existing datasets using the English subsection of OntoNotes 5.0 \citep{recchia_more_2009}. We trained a total of three models using multilingual {\tt xlm-roberta-large}\footnote{https://huggingface.co/xlm-roberta-large} to allow for cross-lingual transfer \citep{conneau_unsupervised_2020}: 1) the first model on 80\% of the preliminary DANSK dataset; 2) the second building on (1) by adding English OntoNotes 5.0 and 3) the third duplicating the 80\% of the preliminary DANSK to match the size of the English OntoNotes 5.0. All three models were validated on the remaining 20\% of the DANSK dataset. The best model (the third, (3)) was then applied to the remaining 15062 texts and discrepancies were manually resolved by the first author.

\paragraph{Resolving remaining inconsistencies}

Because of the large number of annotation reviews, we were able to identify common annotation mistakes. To further enhance the quality of the annotations, all texts were screened for common errors using a list of regex patterns (see \href{https://anonymous.4open.science/r/dansk-3A03} and Appendix A.5.1 upon publication for the full set of regex patterns). 
This resulted in flagged matches in 449 texts which were re-annotated in accordance with the OntoNotes 5.0 extended annotation guidelines \citep{weischedel_ontonotes_2012} and the newly developed Danish Addendum designed to clarify ambiguities and issues specific to Danish texts, as described in the dataset card (Appendix A).

\section{Final dataset: DANSK}
\subsection{DANSK quality assessment}

Average Cohen’s \(\kappa\) scores were calculated on the processed, finalized versions of texts with multiple annotators. All of the non-removed raters’ texts were included, as well as the preliminary version of DANSK with the conflicts resolved. 

\begin{table}[h]
    \centering        


\begin{tabular}{ll}
\toprule

\multicolumn{2}{c}{\textbf{Average Cohen's $\kappa$} (Streamlined)} \\   
\midrule
\midrule
Annotator 1*             & 0.92                           \\         
\rowcolor[HTML]{EFEFEF} 
Annotator 3*             & 0.93                           \\         
Annotator 4*             & 0.93                           \\         
\rowcolor[HTML]{EFEFEF} 
Annotator 5*             & 0.91                           \\         
Annotator 6*             & 0.93                           \\         
\rowcolor[HTML]{EFEFEF} 
Annotator 7*             & 0.93                           \\         
Annotator 8*             & 0.89                           \\         
\rowcolor[HTML]{EFEFEF} 
Annotator 9*             & 0.92                           \\     
\midrule
Preliminary DANSK & 0.92                           \\         
\end{tabular}
        \caption{Table showing the average Cohen's \(\kappa\) scores for each of the non-discarded raters for overlapping data; * indicates that scores are calculated following the automated streamlining process. 
        }
        \label{tab:streamlined_kappa}
\end{table}

As expected, the average scores of the processed texts saw a marked increase, ultimately ranging between 0.93 and 0.89, compared with scores of the original annotated texts which ranged from 0.47 to 0.60 (Table \ref{tab:initial_kappa} and Table \ref{tab:streamlined_kappa}).

\begin{figure}[h]
    \centering
    \includegraphics[scale=.39]{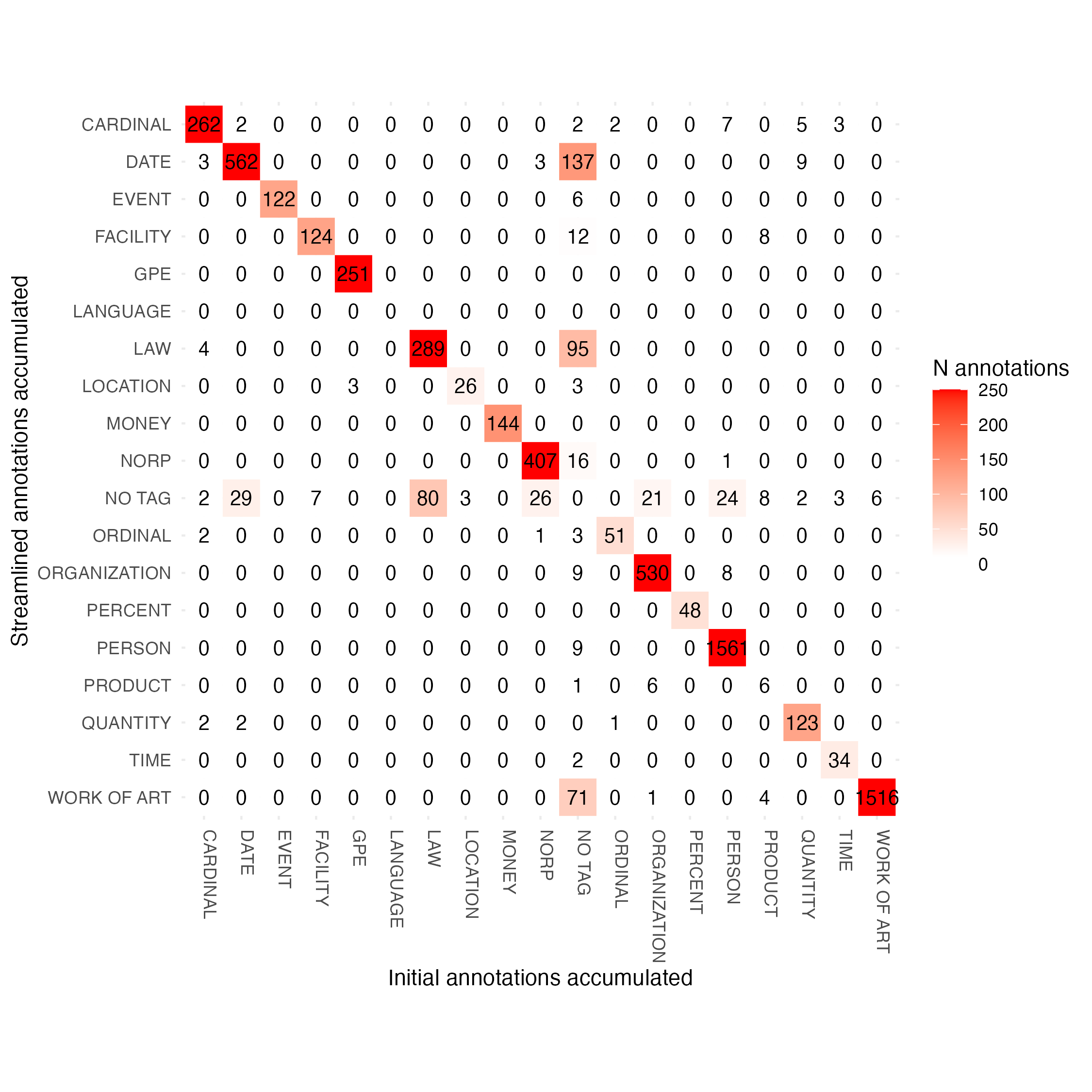}
    \caption{Confusion matrix across annotated tokens before and after the automated streamlining.}
    \label{fig:conf_mtx_streamlined_vs_DANSK}
\end{figure}

To assess which inconsistencies still remained between the DANSK dataset and the raters' annotations, a confusion matrix between the annotations of DANSK and the accumulated annotations of the processed rater texts was assessed. As can be seen in Figure \ref{fig:conf_mtx_streamlined_vs_DANSK}, the majority of  differences are cases in which a token or a span of tokens was considered a named entity by one party, but not by the other. In other words, no unequivocal systematic patterns between a pair of named entities existed.

\subsection{DANSK descriptive statistics} To provide complete transparency about the dataset distributions, descriptive statistics are reported in the dataset card in \href{https://anonymous.4open.science/r/dansk-3A03} and Appendix A with regard to source, domain, and named entities. 

\section{DaCy model curation}

\subsection{Model Specifications}
In order to contribute to Danish NLP with both fine-grained tagging as well as non-domain specific performance, three new models were fine-tuned to the newly developed DANSK dataset. The three models differed in size and included a large, medium, and small model as they were fine-tuned versions of {\tt dfm-encoder-large-v1}\footnote{https://huggingface.co/chcaa/dfm-encoder-large-v1}, {\tt DanskBERT}\footnote{https://huggingface.co/vesteinn/DanskBERT} and {\tt electra-small-nordic}\footnote{https://huggingface.co/jonfd/electra-small-nordic} \citep{snaebjarnarson-etal-2023-transfer}. These models contain 355, 278, and 22 million trainable parameters, respectively. They were chosen based on their ranking among the best-performing Danish language models within their size class, according to the ScandEval benchmark scores current as of the 7th of March, 2023 \citep{nielsen_scandeval_2023}. 

The models were all fine-tuned on the training partition of the DANSK dataset using the Python package \textit{spaCy 3.5.0} \citep{honnibal_spacy_2020}.
The fine-tuning was performed on an NVIDIA T4 GPU through the UCloud interactive HPC system, which is managed by the eScience Center at the University of Southern Denmark. An exhaustive list of all configurations of the system as well as hyperparameter settings is provided in the GitHub repository \footnote{\href{https://anonymous.4open.science/r/DaCy-1BAF}{https://anonymous.4open.science/r/DaCy-1BAF}}.

The three models shared the same hyperparameter settings for the training with the exception that the large model utilized an accumulate gradient of 3. They employed a batch size of 2048 and applied Adam as the optimizer with \(\beta\)1 = 0.9 and \(\beta\)2 = 0.999 and an initial learning rate of 0.0005. It used L2 normalization with weighted decay, \(\alpha\) = 0.01, and gradient clipping with c-parameter = 1.0. For the NER head of the transformer, transition-based parser \citep{goldberg2013training} was used with a hidden width of 64. The models were trained for 20,000 steps with an early stopping patience of 1600. During training the model had a dropout rate of 0.1 and an initial learning rate of 0.0005. 

For the progression of the training loss of the NER head, loss of the transformer, NER performance measured in recall, precision, and F1-score, refer to \href{https://anonymous.4open.science/r/dansk-3A03} and Appendix B.


\subsection{Results}

\begin{figure*}
    \centering
        \includegraphics[width=\textwidth]{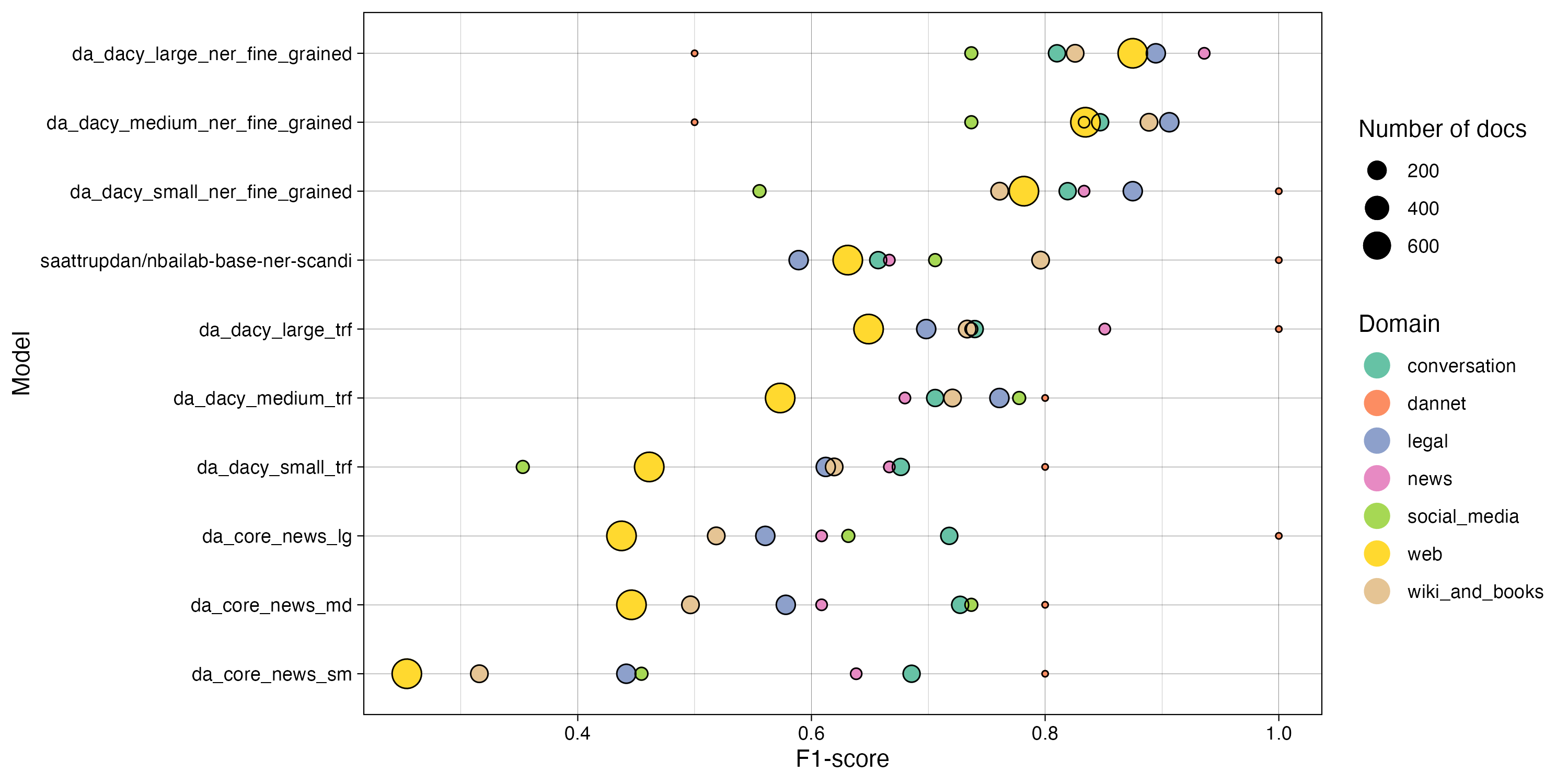}
        \caption{Figure displaying the domain performance in macro F1-scores of the three models on the test partition of DANSK. The size of the circles represents the size of the domains, and thus their relative weighted impact on the overall scores. See Table \ref{tab:ner_domain} for scores.}
        \label{fig:ner_domain}
\end{figure*}

This section presents the results of the performance evaluation.
An overview of the general performance of the three fine-grained models is reported in Table \ref{tab:ner_overall}. Domain-level performance can be seen in Table \ref{tab:ner_domain}. To account for the differences in domain size, Figure \ref{fig:ner_domain} is further included as it adds an additional dimension of information through the depiction of the size of the domains. Insights into performance within named entity categories are provided in Table \ref{tab:ner_tag}.

For full information on distributions for named entities and domains within the partitions, refer to \href{https://anonymous.4open.science/r/dansk-3A03} and Appendix A. 

\begin{table}
    \centering

\begin{tabular}{cccc}
\toprule
\multicolumn{4}{c}{\textbf{DaCy fine-grained model}}                                                              \\ \midrule
\midrule
\cellcolor[HTML]{FFFFFF}                                & \multicolumn{1}{c}{\textbf{Large}}                         & \multicolumn{1}{c}{\textbf{Medium}}                        & \textbf{Small} \\          
\midrule
\rowcolor[HTML]{EFEFEF} 
\multicolumn{1}{l}{\cellcolor[HTML]{EFEFEF}F1-score}  & \multicolumn{1}{c}{\cellcolor[HTML]{EFEFEF}\textbf{0.823}} & \multicolumn{1}{c}{\cellcolor[HTML]{EFEFEF}\textit{0.806}} & 0.776          \\          
\multicolumn{1}{l}{Recall}                            & \multicolumn{1}{c}{\textbf{0.834}}                         & \multicolumn{1}{c}{\textit{0.818}}                         & 0.77           \\          
\rowcolor[HTML]{EFEFEF} 
\multicolumn{1}{l}{\cellcolor[HTML]{EFEFEF}Precision} & \multicolumn{1}{c}{\cellcolor[HTML]{EFEFEF}\textbf{0.813}} & \multicolumn{1}{c}{\cellcolor[HTML]{EFEFEF}\textit{0.794}} & 0.781          \\          
\end{tabular}

    \caption{Table reporting the overall DaCy fine-grained model performances in macro F1-scores. Bold and italics are used to represent the best and second-best scores, respectively.}
    \label{tab:ner_overall}
\end{table}

\begin{table}
    \centering      
    \begin{footnotesize}
\centering
\resizebox{\columnwidth}{!}{%
\begin{tabular}{lccc}
\toprule
\multicolumn{4}{c}{\cellcolor[HTML]{FFFFFF}\textbf{DaCy fine-grained NER}} \\ 
                    \midrule
                    \midrule

\rowcolor[HTML]{FFFFFF} 
\textbf{Named-entity type}                     & \multicolumn{1}{c}{\cellcolor[HTML]{FFFFFF}\textbf{Large}} & \multicolumn{1}{c}{\cellcolor[HTML]{FFFFFF}\textbf{Medium}} & \textbf{Small} \\  
\midrule
\rowcolor[HTML]{EFEFEF} 
CARDINAL                                       & \multicolumn{1}{c}{\cellcolor[HTML]{EFEFEF}\textit{0.87}}       & \multicolumn{1}{c}{\cellcolor[HTML]{EFEFEF}0.78}                 & \textbf{0.89}       \\  
\rowcolor[HTML]{FFFFFF} 
DATE                                           & \multicolumn{1}{c}{\cellcolor[HTML]{FFFFFF}0.85}                & \multicolumn{1}{c}{\cellcolor[HTML]{FFFFFF}\textit{0.86}}        & \textbf{0.87}       \\  
\rowcolor[HTML]{EFEFEF} 
EVENT                                          & \multicolumn{1}{c}{\cellcolor[HTML]{EFEFEF}\textbf{0.61}}       & \multicolumn{1}{c}{\cellcolor[HTML]{EFEFEF}\textit{0.57}}        & 0.4                 \\  
\rowcolor[HTML]{FFFFFF} 
FACILITY                                       & \multicolumn{1}{c}{\cellcolor[HTML]{FFFFFF}\textbf{0.55}}       & \multicolumn{1}{c}{\cellcolor[HTML]{FFFFFF}\textit{0.53}}        & 0.47                \\  
\rowcolor[HTML]{EFEFEF} 
GPE                                            & \multicolumn{1}{c}{\cellcolor[HTML]{EFEFEF}\textbf{0.89}}       & \multicolumn{1}{c}{\cellcolor[HTML]{EFEFEF}\textit{0.84}}        & 0.80                \\  
\rowcolor[HTML]{FFFFFF} 
LANGUAGE                                       & \multicolumn{1}{c}{\cellcolor[HTML]{FFFFFF}\textbf{0.90}}       & \multicolumn{1}{c}{\cellcolor[HTML]{FFFFFF}\textit{0.49}}        & 0.19                \\  
\rowcolor[HTML]{EFEFEF} 
LAW                                            & \multicolumn{1}{c}{\cellcolor[HTML]{EFEFEF}\textbf{0.69}}       & \multicolumn{1}{c}{\cellcolor[HTML]{EFEFEF}\textit{0.63}}        & 0.61                \\  
\rowcolor[HTML]{FFFFFF} 
LOCATION                                       & \multicolumn{1}{c}{\cellcolor[HTML]{FFFFFF}\textit{0.63}}       & \multicolumn{1}{c}{\cellcolor[HTML]{FFFFFF}\textbf{0.74}}        & 0.58                \\  
\rowcolor[HTML]{EFEFEF} 
MONEY                                          & \multicolumn{1}{c}{\cellcolor[HTML]{EFEFEF}\textit{0.99}}       & \multicolumn{1}{c}{\cellcolor[HTML]{EFEFEF}\textbf{1}}           & 0.94                \\  
\rowcolor[HTML]{FFFFFF} 
NORP                                           & \multicolumn{1}{c}{\cellcolor[HTML]{FFFFFF}0.78}                & \multicolumn{1}{c}{\cellcolor[HTML]{FFFFFF}\textbf{0.89}}        & \textit{0.79}       \\  
\rowcolor[HTML]{EFEFEF} 
ORDINAL                                        & \multicolumn{1}{c}{\cellcolor[HTML]{EFEFEF}0.70}                & \multicolumn{1}{c}{\cellcolor[HTML]{EFEFEF}\textit{0.7}}         & \textbf{0.73}       \\  
\rowcolor[HTML]{FFFFFF} 
ORGANIZATION                                   & \multicolumn{1}{c}{\cellcolor[HTML]{FFFFFF}\textbf{0.86}}       & \multicolumn{1}{c}{\cellcolor[HTML]{FFFFFF}\textit{0.85}}        & 0.78                \\  
\rowcolor[HTML]{EFEFEF} 
PERCENT                                        & \multicolumn{1}{c}{\cellcolor[HTML]{EFEFEF}\textit{0.92}}       & \multicolumn{1}{c}{\cellcolor[HTML]{EFEFEF}\textbf{0.96}}        & \textbf{0.96}       \\  
\rowcolor[HTML]{FFFFFF} 
PERSON                                         & \multicolumn{1}{c}{\cellcolor[HTML]{FFFFFF}\textit{0.87}}       & \multicolumn{1}{c}{\cellcolor[HTML]{FFFFFF}\textbf{0.87}}        & 0.83                \\  
\rowcolor[HTML]{EFEFEF} 
PRODUCT                                        & \multicolumn{1}{c}{\cellcolor[HTML]{EFEFEF}\textbf{0.67}}       & \multicolumn{1}{c}{\cellcolor[HTML]{EFEFEF}\textit{0.64}}        & 0.53                \\  
\rowcolor[HTML]{FFFFFF} 
QUANTITY                                       & \multicolumn{1}{c}{\cellcolor[HTML]{FFFFFF}0.39}                & \multicolumn{1}{c}{\cellcolor[HTML]{FFFFFF}\textit{0.65}}        & \textbf{0.71}       \\  
\rowcolor[HTML]{EFEFEF} 
TIME                                           & \multicolumn{1}{c}{\cellcolor[HTML]{EFEFEF}\textit{0.64}}       & \multicolumn{1}{c}{\cellcolor[HTML]{EFEFEF}0.57}                 & \textbf{0.71}       \\  
\rowcolor[HTML]{FFFFFF} 
WORK OF ART                                    & \multicolumn{1}{c}{\cellcolor[HTML]{FFFFFF}\textit{0.49}}       & \multicolumn{1}{c}{\cellcolor[HTML]{FFFFFF}\textbf{0.64}}        & 0.49                \\  
\rowcolor[HTML]{EFEFEF} 
AVERAGE                                        & \multicolumn{1}{c}{\cellcolor[HTML]{EFEFEF}\textbf{0.82}}       & \multicolumn{1}{c}{\cellcolor[HTML]{EFEFEF}\textit{0.81}}        & 0.78                \\  
\end{tabular}%
}
    \end{footnotesize}
    \caption{Table reporting the DaCy fine-grained model performances in F1-scores within each named entity type. Bold and italics are used to represent the best and second-best scores, respectively.}
    \label{tab:ner_tag}
\end{table}

\begin{table}
    \centering        
    \begin{footnotesize}

\centering
\begin{tabular}{lccc}
\toprule
\multicolumn{4}{c}{\textbf{DaCy fine-grained model}}                                                           \\ \midrule
\midrule
\rowcolor[HTML]{EFEFEF} \textbf{Domain}      & \textbf{Large} & \textbf{Medium} & \textbf{Small} \\   
\midrule
All domains  & \textbf{0.82}       & \textit{0.81}        & 0.78                \\   
\rowcolor[HTML]{EFEFEF} 
Conversation         & \textit{0.80}       & 0.72                 & \textbf{0.82}       \\   
Dannet               & \textit{0.75}       & 0.667                & \textbf{1}          \\   
\rowcolor[HTML]{EFEFEF} 
Legal                & \textit{0.85}       & \textit{0.85}        & \textbf{0.87}       \\   
News                 & \textit{0.84}       & 0.76                 & \textbf{0.86}       \\   
\rowcolor[HTML]{EFEFEF} 
Social Media         & 0.79                & \textbf{0.85}        & \textit{0.8}        \\   
Web                  & \textbf{0.83}       & \textit{0.80}        & 0.76                \\   
\rowcolor[HTML]{EFEFEF} 
Wiki and Books       & \textit{0.78}       & \textbf{0.84}        & 0.71                \\   
\end{tabular}

  \end{footnotesize}
    \caption{Table reporting the DaCy fine-grained model performances in macro F1-scores within each domain. Bold and italics are used to represent the best and second-best scores, respectively.}
    \label{tab:ner_domain}
\end{table}

\section{Model generalizability}
\subsection{Methods}
\subsubsection{Models}
To assess whether there exists a generalizability issue for Danish language models, a number of SOTA models were chosen for evaluation on the test partition of the newly developed DANSK dataset. 
The field of Danish NLP and NER is evolving rapidly, making it hard to establish an overview of the most important models for Danish NER. 
However, the models for the evaluation were chosen on the basis of two factors; namely prominence of use, and performance. The latter was gauged on the basis of ScandEval, the NLU framework for benchmarking \cite{nielsen_scandeval_2023}.  

At the time of the model search, the model {\tt saattrupdan/nbailab-base-ner-scandi}
\footnote{https://huggingface.co/saattrupdan/nbailab-base-ner-scandi} ranked amongst the best-performing models for Danish (and scandinavian) NER.\footnote{https://paperswithcode.com/sota/named-entity-recognition-on-dane} It was trained on the combined dataset of DaNE, NorNE, SUC 3.0, and the Icelandic and Faroese part of the WikiANN \citep{hvingelby2020dane, gustafson-capkova_manual_2006, ejerhed_linguistic_1992, jorgensen_norne_2019, pan_cross-lingual_2017}.
Because of the wide palette of different datasets, texts from more domains are represented. It was thus conjectured that the model might not suffer from the generalizability issues outlined in the introduction section of the paper.

Apart from this model, the three v0.1.0 DaCy models large, medium, and small were also included. Note that these are the existing non-fine-grained models that were already in DaCy prior to the development of the fine-grained DaCy models presented in this paper. The models are fine-tuned versions of 1) Danish Ælæctra\footnote{https://huggingface.co/Maltehb/aelaectra-danish-electra-small-cased}, Danish BERT\footnote{https://huggingface.co/Maltehb/danish-bert-botxo}, and the XLM-R \cite{conneau_unsupervised_2020}. 
The models are fine-tuned on DaNE \cite{hvingelby2020dane} and DDT \cite{johannsen2015universal} for multitask prediction for multiple task including named-entity recognition and at the time of publication achieved state-of-the-art performance for Danish NER \citep{enevoldsen_dacy_2021}.

 We also include the NLP framework \textit{spaCy} (Explosion AI, Berlin, Germany), to explore the generalization of production systems. SpaCy features three Danish models (small, medium, and large\footnote{Note that a model size of spaCy are not comparable to model sizes of transformer encoders}) which similarly to the DaCy models are multi-task models with NER capabilities. Although spaCy also includes a Danish transformer model, it was not incorporated in the generalizability analysis. The reason for this is that DaCy medium v.0.1.0 is already included and the two models are almost identical. Both are based on the model {\tt Maltehb/danish-bert-botxo}\footnote{https://huggingface.co/Maltehb/danish-bert-botxo} and fine-tuned on DaNE, and thus only deviate on minor differences in hyperparameter settings. 

In summary, the models included in the final evaluation were:

\begin{footnotesize}

\begin{enumerate}
    \itemsep-0.25em 
    \item {\tt Base-ner-scandi} \newline ({\tt nbailab-base-ner-scandi})
    \item {\tt spaCy large} \newline({\tt da\_core\_news\_lg v. 3.5.0})
    \item {\tt spaCy medium} \newline({\tt da\_core\_news\_md v. 3.5.0})
    \item {\tt spaCy small} \newline({\tt da\_core\_news\_sm v. 3.5.0})
    \item {\tt DaCy (fine-grained) large} ({\tt da\_dacy\_large\_trf-0.1.0})
    \item {\tt DaCy (fine-grained) medium} ({\tt da\_dacy\_medium\_trf-0.1.0})
    \item {\tt DaCy (fine-grained) small} ({\tt da\_dacy\_small\_trf-0.1.0})
    
\end{enumerate}
\end{footnotesize}

\subsubsection{Named Entity Label Transfer}\label{subsec:ne_label_transfer}
A fine-grained NER dataset with 18 labels following the OntoNotes guidelines has not been publicly available for Danish until now. The aforementioned models have thus only been fine-tuned to the classic, more coarse-grained DaNE dataset that follows the CoNLL-2003 named entity annotation scheme \citep{sang_introduction_2003, hvingelby_dane_2020}. This includes the four named entity types PER (person), LOC (location), ORG (organization), and MISC (miscellaneous). This annotation scheme is radically different from the DANSK annotations that match the OntoNotes 5.0 standards. To enable an evaluation of the models, the DANSK named entity labels were coerced into the CoNLL-2003 standard in order to match the nature of the models, and specifically to assist us in highlighting performance disparities across out-of-distribution domains, such as "SoMe" and "Legal", which are new in the release of DaNSK. 

As the description of both ORG and PER in CoNLL-2003 largely matches that of the extended OntoNotes, these named entity types could be used in the evaluation with a 1-to-1 mapping without further handling. However, in CoNLL-2003, LOC includes cities, roads, mountains, abstract places, specific buildings, and meeting points \citep{hvingelby_dane_2020, sang_introduction_2003}. As the extended OntoNotes guidelines use both GPE and LOCATION,
DANSK GPE annotations were mapped to LOC in an attempt to make the test more accurate. Predictions for the CoNLL-2003 MISC category, intended for names not captured by other categories (e.g. events and adjectives such as "\textit{2004 World Cup}" and "\textit{Italian}"), were excluded. 


\subsubsection{Evaluation}
SOTA models were evaluated using macro average F1-statistics at a general level, a domain level, and finally F1-scores at the level of named entity types.

\subsection{Results}

Table \ref{tab:sota_overall} provides an overview of macro span-F1-scores as well as recall and precision statistics. The performance across domains and across named entity types are reported in Table \ref{tab:sota_domain} and Table \ref{tab:sota_tags}. 


\begin{table}[ht]
    \centering        
   \begin{footnotesize}
\centering
\begin{tabular}{|c|c|c|c|}
\hline
\rowcolor[HTML]{EFEFEF} 
\textbf{Model}  & \textbf{F1}                 & \textbf{Recall}             & \textbf{Precision}          \\ \hline
\rowcolor[HTML]{FFFFFF} 
Base-ner-scandi & \textit{0.64}               & 0.59                        & \textbf{0.70}               \\ \hline
\rowcolor[HTML]{EFEFEF} 
DaCy large     & \textbf{0.68}               & \textbf{0.67}               & \textit{0.69}               \\ \hline
\rowcolor[HTML]{FFFFFF} 
DaCy medium    & 0.63                        & \textit{0.64}               & 0.61                        \\ \hline
\rowcolor[HTML]{EFEFEF} 
DaCy small     & 0.51                        & 0.48                        & 0.56                        \\ \hline
\rowcolor[HTML]{FFFFFF} 
spaCy large     & 0.49                        & 0.45                        & 0.53                        \\ \hline
\rowcolor[HTML]{EFEFEF} 
spaCy medium    & 0.49                        & 0.47                        & 0.52                        \\ \hline
\rowcolor[HTML]{FFFFFF} 
spaCy small     & {\color[HTML]{000000} 0.32} & {\color[HTML]{000000} 0.32} & {\color[HTML]{000000} 0.32} \\ \hline
\end{tabular}
   \end{footnotesize}
    \caption{Table showing the overall performance in macro F1-scores on the DANSK test set. Bold and italic represent the best and next best scores.}
    \label{tab:sota_overall}
\end{table}


\begin{table*}
    \centering

\centering
\begin{tabular}{l|cccccccc|}
\hline

\textbf{Model}  & Across & Conversational  & Legal & News & SoMe & Web & Wiki \\ 
\hline
\hline
base-ner-scandi & \textit{0.64}   & 0.66                        & 0.59           & \textit{0.67} & 0.71                  & \textit{0.63} & \textbf{0.80} \\ 

DaCy Large      & \textbf{0.68}   & \textbf{0.74}              & \textit{0.70}  & \textbf{0.85} & \textit{0.74}         & \textbf{0.65} & \textit{0.73} \\ 

DaCy Medium    & 0.63            & 0.71                      & \textbf{0.76}  & 0.68          & \textbf{0.78}         & 0.57          & 0.72          \\ 

DaCy Small     & 0.51            & 0.68                     & 0.61           & \textit{0.67} & 0.35                  & 0.46          & 0.62          \\ 

spaCy Large      & 0.49            & 0.72                        & 0.56           & 0.61          & 0.63                  & 0.44          & 0.52          \\ 

spaCy Medium   & 0.49            & \textit{0.73}             & 0.58           & 0.61          & \textit{0.74}         & 0.45          & 0.50          \\ 

spaCy small     & 0.32            & 0.69                    & 0.44           & 0.64          & 0.46                  & 0.25          & 0.32          \\ 
\hline
\end{tabular}

    \caption{Table showing the domain performances in macro F1-scores of the models on the DANSK test set. Bold and italic represent the best and next best scores.}
    \label{tab:sota_domain}
\end{table*}


\section{Discussion}


\subsection{DANSK dataset}

The DANSK dataset enhances Danish NER by focusing on fine-grained named entity labels and diverse domains like conversations, legal matters, and web sources, but omits some domains in DaNE, such as magazines \citep{norling-christensen_corpus_1998, hvingelby_dane_2020}. Entity distribution varies, influencing model performance for specific types.

DANSK's quality was benchmarked using models trained on different OntoNotes 5.0 annotated datasets \cite{luoma_fine-grained_2021}. Despite the dataset size disparity, performances for English and Finnish models were between F1-scores of .89 and .93 \citep{luoma_fine-grained_2021, li_unified_2022}, notably higher than DANSK. Given the smaller size of DANSK (15062 texts) compared to English OntoNotes (600000 texts) \citep{weischedel_ontonotes_2012}, performance for models trained on DANSK is expectedly lower, irrespective of annotation quality \citep{russakovsky_imagenet_2015}.

Annotation quality issues were tackled, improving Cohen's $\kappa$ values from $\sim$0.5 to $\sim$0.9 (Table \ref{tab:initial_kappa} and Table \ref{tab:streamlined_kappa}). Initial difficulties arose from suboptimal sampling from DAGW and insufficient annotator training. Future improvements include initial quality screening and comprehensive training with the OntoNotes 5.0 annotation scheme \citep{plank_problem_2022, uma_semeval-2021_2021}. In the release of the DANSK dataset, we include raw (per annotator) annotations to allow for transparency and further analysis of annotator disagreement. 


\subsection{DaCy models}

New fine-grained models of varying sizes attained macro F1-scores of 0.82, 0.81, and 0.78 respectively. Larger models generally performed better as would be expected. However, due to DANSK's domain imbalance, these scores should be treated carefully. Domains like web, conversation, and legal heavily influenced the F1-scores due to their larger text volume. Performance comparisons are based on OntoNotes 5.0 standard datasets due to the unique annotation scheme of DANSK.

Minor performance variation was found within each domain. The small models excelled in underrepresented domains like news, possibly leading to volatile results. Legal texts were easiest to classify with F1-scores of 0.85 and 0.87.

Classification performance varied with named entity types. Facilities, artworks, and quantities were difficult to predict, whereas entities like money, dates, percentages, GPEs, organizations, and cardinals were easier to classify. This can be attributed to the quantity and context of named entities in the training data. Some entity types might appear in similar contexts or have similar structures, hence easier to distinguish. Variance in performance may arise from differences in text quality and context. Given the observed performance differences across domains and named entity types, it's crucial to understand the strengths and limitations of the new models within the DaCy framework.

\begin{table*}[ht!]
    \centering       
    \begin{footnotesize}
        \begin{tabular}{lcccc}
\toprule
\textbf{Model} & \textbf{Average F1} & \textbf{Person F1} & \textbf{Organization F1} & \textbf{Location F1} \\

\midrule
\midrule
\rowcolor[HTML]{EFEFEF} DaCy large & 0.67 (0.62, 0.72) & 0.74 (0.67, 0.80) & 0.50 (0.43, 0.57) & 0.80 (0.73, 0.86) \\
DaCy medium & 0.56 (0.49, 0.60) & 0.62 (0.54, 0.68) & 0.40 (0.32, 0.47) & 0.66 (0.53, 0.75) \\
\rowcolor[HTML]{EFEFEF} DaCy small  & 0.55 (0.50, 0.59) & 0.64 (0.56, 0.71) & 0.38 (0.31, 0.46) & 0.65 (0.56, 0.72) \\
base-ner-scandi & 0.64 (0.57, 0.69) & 0.69 (0.62, 0.77) & 0.49 (0.38, 0.59) & 0.72 (0.58, 0.81) \\
\rowcolor[HTML]{EFEFEF} SpaCy large & 0.51 (0.43, 0.56) & 0.60 (0.52, 0.68) & 0.33 (0.24, 0.42) & 0.61 (0.46, 0.71) \\
SpaCy Medium & 0.50 (0.44, 0.55) & 0.59 (0.51, 0.65) & 0.32 (0.26, 0.41) & 0.62 (0.48, 0.72) \\
\rowcolor[HTML]{EFEFEF} SpaCy Small & 0.34 (0.30, 0.40) & 0.36 (0.29, 0.43) & 0.22 (0.16, 0.29) & 0.46 (0.35, 0.55) \\
\midrule
DaCy fine-grained large (ours) & \textbf{0.85} (0.81, 0.88) & \textit{0.86} (0.80, 0.90) & \textit{0.79} (0.73, 0.85) & \textbf{0.93} (0.89, 0.96) \\
\rowcolor[HTML]{EFEFEF} DaCy fine-grained medium (ours) & \textbf{0.85} (0.81, 0.88) & 0.85 (0.79, 0.90) & \textbf{0.80} (0.76, 0.85) & \textit{0.91} (0.86, 0.96) \\
DaCy fine-grained small (ours) & \textit{0.83} (0.80, 0.86) & \textbf{0.87} (0.82, 0.92) & \textit{0.79} (0.74, 0.83) & 0.85 (0.78, 0.92) \\
\bottomrule
\end{tabular}
    \end{footnotesize}
    \caption{Table showing the performance in F1-scores within each of the named entity classes on the DaNSK test set. Bold and italic represent the best and next best scores. Scores are bootstrapped on the documents level and shows the mean the 95\% confidence interval in showed in the parentheses.}
    \label{tab:sota_tags}
\end{table*}

\subsection{SOTA models and generalizability}

The new fine-grained DaCy models demonstrate higher performance on the DANSK dataset, compared to existing SOTA models (refer to Tables \ref{tab:sota_overall} and \ref{tab:ner_overall}). However, due to annotation scheme discrepancies, a direct comparison is challenging.

Performance analysis is two-fold: evaluation across domains for each model, and comparison between models, both following the CoNLL-2003 annotation scheme.

Significant domain performance disparities were observed (see Table \ref{tab:sota_domain}). For instance, {\tt base-ner-scandi} scored F1-scores of 0.59 and 0.8 for legal and Wikipedia texts, respectively. Actual model accuracy may vary by domain, contrary to performance reported on DaNE. The models performed best on conversation and news texts, with web and wiki sources performing poorly.

Larger models generally outperformed smaller models, with {\tt base-ner-scandi} and {\tt DaCy large} performing best, with across-domain F1-scores of  0.64 and 0.68 respectively. The DaCy models, easily accessible via the DaCy framework, performed comparably or better than the {\tt base-ner-scandi} model, hence DaCy is the preferred library for Danish NER.

Table \ref{tab:sota_tags} shows the performance of models within each non-fine-grained named entity class (CoNLL-2003) on the DaNSK test set, and includes scores for the previously best-performing non-fine-grained DaCy models (0.2.0). The release of fine-grained NER DaCy models (0.1.0) represents a significant performance improvement, from an overall average F1-score of 0.67 for DaCy Large (0.2.0) versus 0.85 for DaCy fine-grained large (0.1.0).  


\section{Conclusion}

Danish NER suffers from limited dataset availability, lack of cross-validation, domain-specific evaluations, and fine-grained NER annotations. This paper introduces DANSK, a high-granularity named entity dataset for within-domain evaluation, DaCy 2.6.0 with three generalizable, fine-grained models, and an evaluation of contemporary Danish models. DANSK, annotated following OntoNotes 5.0 and including metadata on text origin, facilitates across-domain evaluations. However, observed performance still falls short of what is seen among higher-resourced languages. DaCy models, trained on DANSK, achieve up to 0.82 macro F1-score on fine-grained NER across 18 categories. While work remains to be done to augment the size and quality of fine-gained named entity annotation in Danish, the release of DANSK and DaCy will assist in addressing generalizability issues in the field. 

\section{Ethics statement}

Ethics Statement
Our dataset is constructed based on the public dataset The Danish Gigaword corpus, which followed ethical practices in its composition. For spoken conversations, participants
agreed on releasing anonymized transcripts of their
conversations. Social media data only includes publicly available Tweets. Because distribution of this part of the dataset is through Tweet IDs and requires rehydration, any Tweets subsequently removed by the user are no longer included. 

10 native Danish speakers enrolled in the English Linguistics Master's program were recruited as annotators through announcements in classrooms. This degree program was chosen because students receive relevant training in general linguistics, including syntactic analysis. We employed a larger group of students to adhere to institutional limitations on the number of hours student workers can have. The demographic bias of our annotators (nine female, one male) reflects the demographics of this MA program. Annotators worked 10 hours/week for six weeks from October 11,
2021, to November 22, 2021. Their annotation tasks included
part-of-speech tagging, dependency parsing, and NER annotation. Annotators were compensated at the standard rate for students, as determined by the collective agreement of the Danish Ministry of Finance and the Central Organization of Teachers and the
CO10 Central Organization of 2010 (the CO10 joint agreement), which is 140DKK/hour.

We are committed to full transparency and replicability in our release of DaNSK. Following work by \citet{mitchell2019model} and \cite{gebru2021datasheets}, we provide a dataset card for DANSK following the format proposed in \citet{lhoest2021datasets}, which can be accessed here: \href{https://anonymous.4open.science/r/dansk-3A03}{https://anonymous.4open.science/r/dansk-3A03}. The datasetcard and additional supporting information about the language resource will also be included in Appendices upon publication.

\newpage

\section{Bibliographical References}\label{sec:reference}

\bibliographystyle{lrec-coling2024-natbib}
\bibliography{references}

\end{document}